\documentclass[10pt,twocolumn,letterpaper]{article}

\usepackage[pagenumbers]{cvpr} 

\usepackage[dvipsnames]{xcolor}

\definecolor{cvprblue}{rgb}{0.21,0.49,0.74}
\usepackage[pagebackref,breaklinks,colorlinks,citecolor=cvprblue]{hyperref}
\usepackage{multirow}
\usepackage{algorithm}
\usepackage{algorithmicx}
\usepackage{algpseudocode}
\usepackage{balance}

\def\paperID{8619} 
\def\confName{CVPR}
\def\confYear{2024}

\title{Point2CAD: Reverse Engineering CAD Models from 3D Point Clouds}

\author{Yujia Liu\\
ETH Z\"{u}rich\\
\and
Anton Obukhov\\
ETH Z\"{u}rich\\
\and
Jan Dirk Wegner\\
University of Z\"{u}rich\\
\and
Konrad Schindler\\
ETH Z\"{u}rich\\
}

\begin{document}
\twocolumn[{%
  \renewcommand\twocolumn[1][]{#1}%
\maketitle
\begin{center}
  \newcommand{\teaserwidth}{1.0\textwidth}
  \centerline{
    \includegraphics[width=1.0\linewidth,trim={0 0 0 3cm},clip]{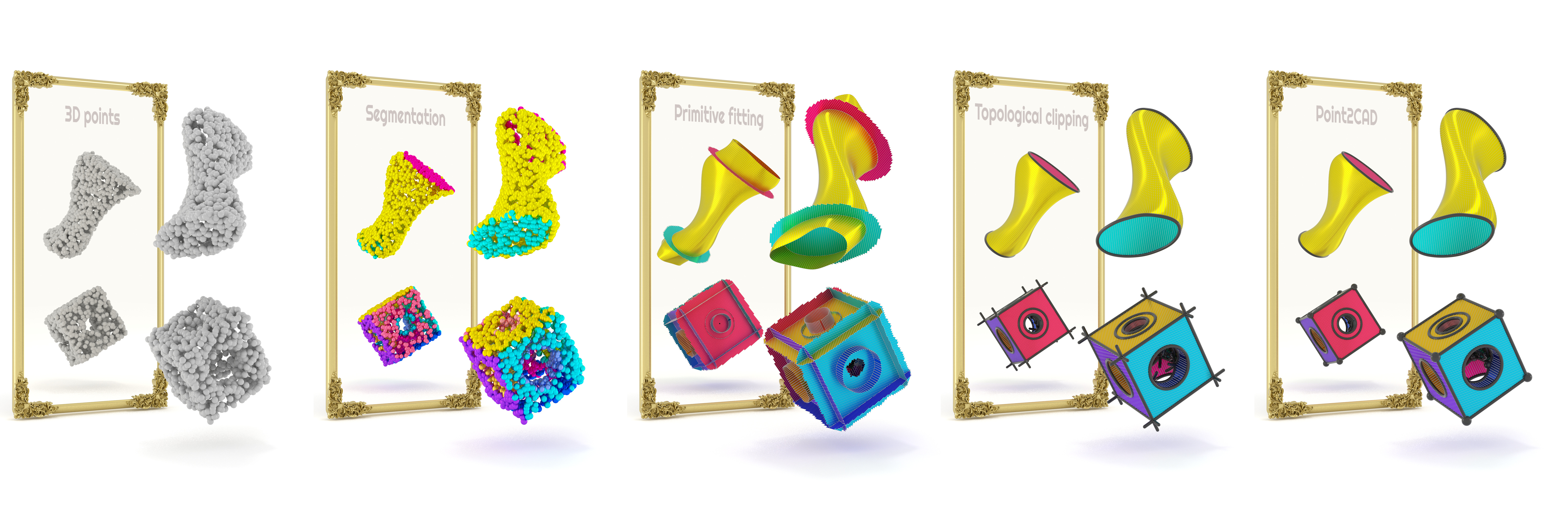}
  }
  \captionsetup{type=figure}
  \captionof{figure}{
    \textbf{Point2CAD} is a pipeline for reconstructing complex CAD models, including freeform surfaces, from 3D point clouds. 
    A raw point cloud is segmented into clusters corresponding to CAD faces, and each face is fitted either with a geometric primitive or with a parametric freeform surface, using a novel, implicit neural representation. Due to the analytic representation the surfaces can be extended and intersected, such that
    topology emerges, which is then used to clip the surface primitives.
  }
  \label{fig:teaser}
 \end{center}%
}]

\begin{abstract}
\label{sec:abstract}
Computer-Aided Design (CAD) model reconstruction from point clouds is an important problem at the intersection of computer vision, graphics, and machine learning; it saves the designer significant time when iterating on in-the-wild objects.
Recent advancements in this direction achieve relatively reliable semantic segmentation but still struggle to produce an adequate topology of the CAD model. 
In this work, we analyze the current state of the art for that ill-posed task and identify shortcomings of existing methods.
We propose a hybrid analytic-neural reconstruction scheme that bridges the gap between segmented point clouds and structured CAD models and can be readily combined with different segmentation backbones.
Moreover, to power the surface fitting stage, we propose a novel implicit neural representation of freeform surfaces, driving up the performance of our overall CAD reconstruction scheme.
We extensively evaluate our method on the popular ABC benchmark of CAD models and set a new state-of-the-art for that dataset.
Project page: \href{https://www.obukhov.ai/point2cad}{https://www.obukhov.ai/point2cad}.

\end{abstract}    
\section{Introduction}
\label{sec:intro}

The task of reverse engineering CAD models from 3D point clouds has gained increasing attention in recent years due to the rapid development of 3D scanning technologies. 
Reverse engineering involves transforming a physical object into a digital model in an editing-friendly format, which can be used for analysis, visualization, and manufacturing. 

Most approaches to reverse engineering CAD models follow a typical sequence of steps. First, data capture and pre-processing are carried out to obtain point cloud data sampled densely from the object's surface. Subsequently, the point cloud is segmented to identify individual regions of interest. The regions are then classified to differentiate between different types of surfaces and features. Analytical representations are generated based on the identified regions, followed by the reconstruction of the model's topology. 
Commercial software systems offer interactive tools and functions for each step, such as a dedicated segmentation or shape-fitting tool.
Similarly, research that aims to automate the process has focused chiefly on individual steps, whereas little work covers the complete workflow of CAD model reconstruction. 
Furthermore, most existing work is either limited to case studies of a few models or is still quite far from producing satisfactory results.

In particular, reconstruction with simple parametric surface primitives
has been studied extensively. 
For such surfaces, both fitting and computing intersections are comparatively straight-forward, and with sufficiently clean point clouds, a boundary representation ($B$-rep) can be derived with classical geometric methods~\cite{wang2012framework, alai2013review, durupt20083d}. 
On the contrary, while there are several works about fitting freeform surfaces, often with various flavors of splines, they mostly stop short of assembling individual surfaces into complete $B$-reps~\cite{gao2019deepspline, sharma2020parsenet, yan2021hpnet}. 
In summary, geometrically complex free-form surfaces and topologically complex assemblies of simple surfaces have mostly been studied separately.
The present work addresses this gap by proposing \textbf{Point2CAD}, a method that recovers complete CAD models, including free-form surfaces, edges, and corners. Refer to Sec.~\ref{sec:notation} of sup.mat. for notation.\footnote{A note on terminology: many different terms have been used to denote the topological elements of CAD models, see list in the supplementary material. We use the terms \emph{surface} (2D), \emph{edge} (1D) and \emph{corner} (0D).}

Our proposed pipeline for CAD model reconstruction from point clouds comprises several steps. 
First, a pre-trained neural network segments the point cloud into clusters corresponding to distinct surfaces. 
Second, surfaces are fitted to the clusters, which involves both basic primitive fitting and a novel neural fitting scheme for freeform surfaces. 
Third, adjacent surfaces are intersected to recover edges, and adjacent edges are further intersected to recover corners, thus obtaining a full $B$-rep.
Taken together, these steps form a comprehensive and versatile pipeline for reverse engineering point clouds into CAD models; see Fig.~\ref{fig:teaser}.
By combining modern, learning-based segmentation backbones with classical geometric primitive fitting and with recent neural field methods, we get the best of both worlds and obtain a reconstruction pipeline that sets a new state of the art on the large and diverse ABC benchmark~\cite{koch2019abc}.
Our contributions can be summarized as follows:
\begin{itemize}[leftmargin=\parindent] 
\item We propose Point2CAD, a comprehensive workflow for reconstructing complex free-form parametric CAD models from raw 3D point clouds;
\item To power the surface fitting stage, we propose a novel neural representation of freeform surfaces;
\item We demonstrate the superior quantitative and qualitative performance on the challenging ABC dataset.
 \end{itemize}

\section{Related work}
\label{sec:related-work}

Our work aims to recover CAD models from raw point clouds. 
Towards this goal, we draw inspiration from several domains of computer vision and 3D data processing.

\noindent\textbf{3D point cloud segmentation.} Many studies segment point clouds by assigning each point a class or instance label. Neural architectures to learn feature embeddings of point clouds include: 
unordered sets of neighbors~\cite{qi2017pointnet, qi2017pointnet++}, 
graph convolutions~\cite{wang2019dynamic, li2019deepgcns}, 
point convolutions~\cite{li2018pointcnn, hua2018pointwise}, 
point-voxel learning~\cite{rethage2018fully, liu2019point}, and transformers~\cite{zhao2021point, guo2021pct}. 
Several authors use point-wise labeling as a mechanism to structure point clouds into geometric primitives~\cite{he2018survey}.
\Eg, PrimitiveNet~\cite{huang2021primitivenet} segments point clouds into primitives via local embedding and adversarial learning. 
In \cite{loizou2020learning} graph convolution is used to detect surface boundaries in 3D point clouds.

\noindent\textbf{Primitive fitting} \cite{farin2002curves, kaiser2019survey} is a common way to abstract 3D point clouds into CAD-like parametric elements.
Classical solutions rely on robust statistical procedures like the Hough transform to recognize geometric primitives in point clouds despite the noise and missing data~\cite{romanengo2023recognising}.
Several learning-based approaches have also been proposed to fit geometric primitives to point clouds. 
SPFN~\cite{li2019supervised} is trained to detect varying numbers of primitives at different scales, supervised with ground truth surfaces and their membership in the primitives. 
ParSeNet~\cite{sharma2020parsenet} finds parametric surfaces in point clouds, including basic geometric primitives as well as B-spline surfaces, but does not put much effort into connecting them. 
HPNet~\cite{yan2021hpnet} concentrates on partitioning the point cloud into segments using semantic and spectral features and edge information in the form of an adjacency matrix but does not fit actual primitives.
Saporta et al.~\cite{saporta2022unsupervised} proposed an unsupervised, recursive neural architecture that learns to fit geometric primitives to 3D points, while Vasu et al.~\cite{vasu2021hybridsdf} develop a hybrid representation that combines explicit geometric shapes with signed distance functions to represent and manipulate regular and freeform 3D shapes. 
Besides conventional geometric primitives, parametric freeform surfaces are an important building block of CAD systems. DeepSpline~\cite{gao2019deepspline} utilizes a hierarchical RNN to reconstruct 2D spline curves, in conjunction with an unsupervised learning approach to recover surfaces of revolution and extrusion.

\noindent\textbf{CAD representations.}
BRepNet~\cite{lambourne2021brepnet} is a neural network architecture that operates directly on $B$-rep data structures. 
UV-Net~\cite{jayaraman2021uv} introduces another neural architecture that operates on $B$-rep data, without fitting to points.
JoinABLe~\cite{willis2022joinable} starts from CAD primitives and learns to assemble them and form joints, using weak supervision available in standard CAD files. 
Another popular CAD representation is constructive solid geometry (CSG). 
CAPRI-Net~\cite{yu2022capri} learns to fit a collection of quadric 3D primitives, and an associated CSG-tree, while CSGNet~\cite{sharma2020neural} learns to generate CSG programs for given 2D or 3D shapes, using a convolutional encoder and a recurrent decoder. 

\noindent\textbf{Generic CAD modeling}.
Reconstructing a complete CAD model~\cite{mozaffar2021mechanistic, buonamici2018reverse} is challenging, as the output should include both the geometric surface shapes and their connectivity.
Traditional methods for reverse engineering often employ iterative RANSAC estimation to extract primitive shapes from point clouds, \eg,~\cite{liu2020adaptive}. 
Benko et al.~\cite{benkHo2001algorithms} outline a hand-engineered, sequential procedure to reconstruct CAD models via segmentation, surface and topology fitting, and blending. 
Recent, learning-based approaches in the same spirit include Point2Cyl~\cite{uy2022point2cyl}, a neural network that transforms a 3D point cloud into a set of extrusion cylinders by predicting point segmentation, base labels, normals, and extrusion parameters in closed form.
ComplexGen~\cite{guo2022complexgen} reconstructs CAD models by autoregressively detecting geometric primitives. The model consists of a sparse CNN encoder, three transformer decoders for geometric primitives and topology, and post-processing with global optimization to refine the rather rough transformer predictions.

\noindent\textbf{Manifold learning} refers to a family of methods that aim to discover a low-dimensional manifold underlying a higher-dimensional, possibly noisy, data set. Manifold learning methods can be understood as non-linear extensions of PCA~\cite{pca} attempting to ``flatten'' curved manifolds. Most methods focus on visualization or fidelity to particular input points, by explicitly using those points to construct the mapping~\cite{locallylinearembedding,isomaps,umap}.
Autoencoders provide a natural way to not only project data non-linearly to a lower-dimensional latent space but also to decode back from the latent space to the original data space~\cite{kramer1991nonlinear}. 
We extend this approach with the recent findings about implicit neural representations to design a fitting method for freeform surfaces.

\noindent\textbf{Implicit Neural Representations (INR)} are a framework to encode an arbitrary function, often observed in the form of sparse samples, into the weights of a neural network.
Recent research about neural rendering has led to several useful and transferrable insights about implicit neural fields, for instance, the importance of positional encoding~\cite{nerf} and new (e.g., sinusoidal) activation functions~\cite{siren}. These findings form the basis for our freeform surface fitting method.

\begin{figure}[t]
\begin{center}
   \includegraphics[width=1.0\linewidth]{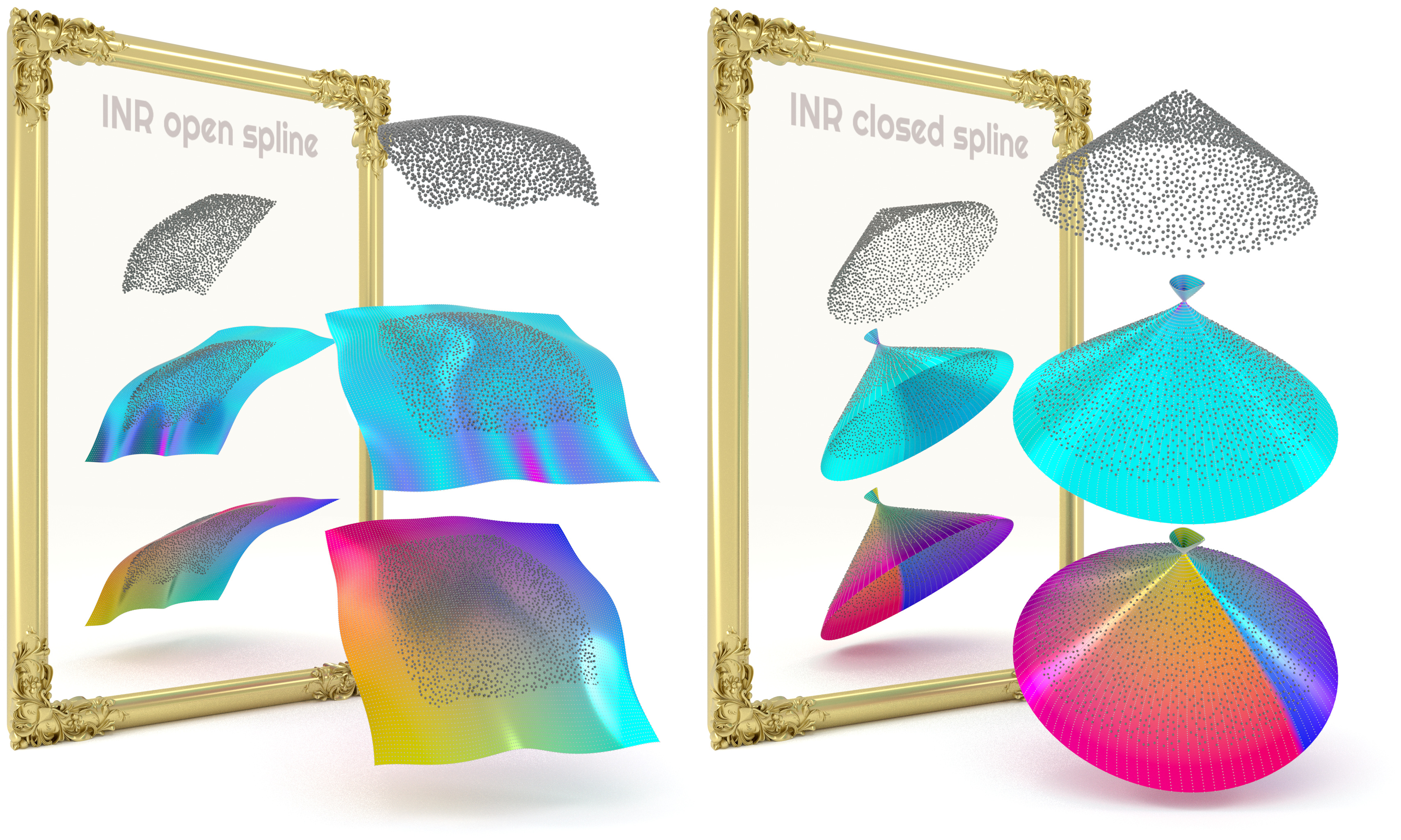}
\end{center}
    \caption{
    Visualization of open (\textbf{left}) and closed (\textbf{right}) surface fitting with the proposed INR. 
    \textbf{Top}: point cloud sampled from a freeform surface; 
    \textbf{middle}: INR fit, showing \textcolor{magenta}{high} or \textcolor{cyan}{low} principal curvature; 
    \textbf{bottom}: same fit with color indicating $uv$ coordinates. 
    Our INR reproduces curvature where necessary and smoothly extrapolates without oscillations to facilitate subsequent steps.
        }
\label{fig:inr:good}
\end{figure}

\begin{figure*}[t!]
\begin{center}
   \includegraphics[width=0.24\linewidth]{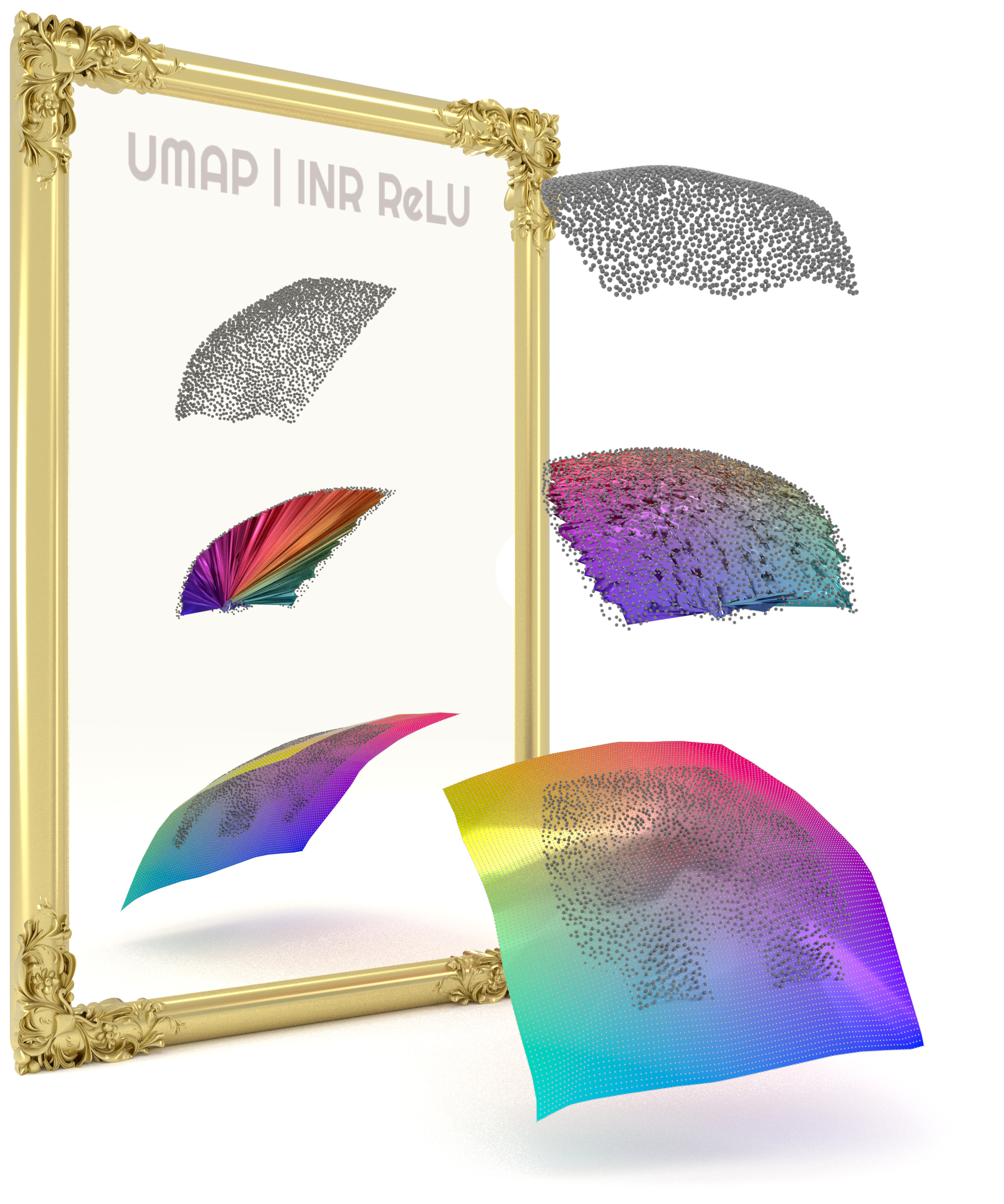}
      \hfill
   \includegraphics[width=0.24\linewidth]{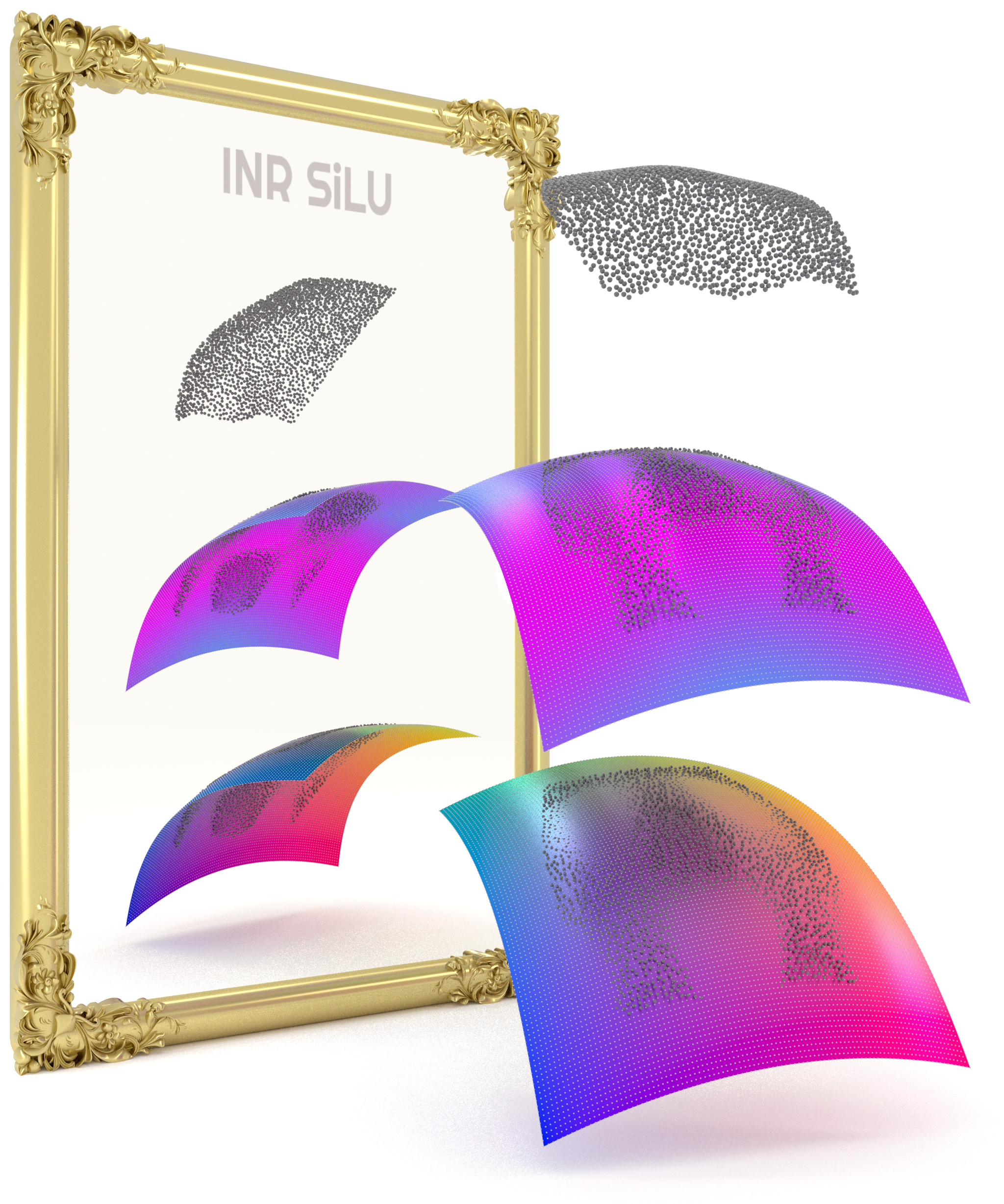}
      \hfill
   \includegraphics[width=0.24\linewidth]{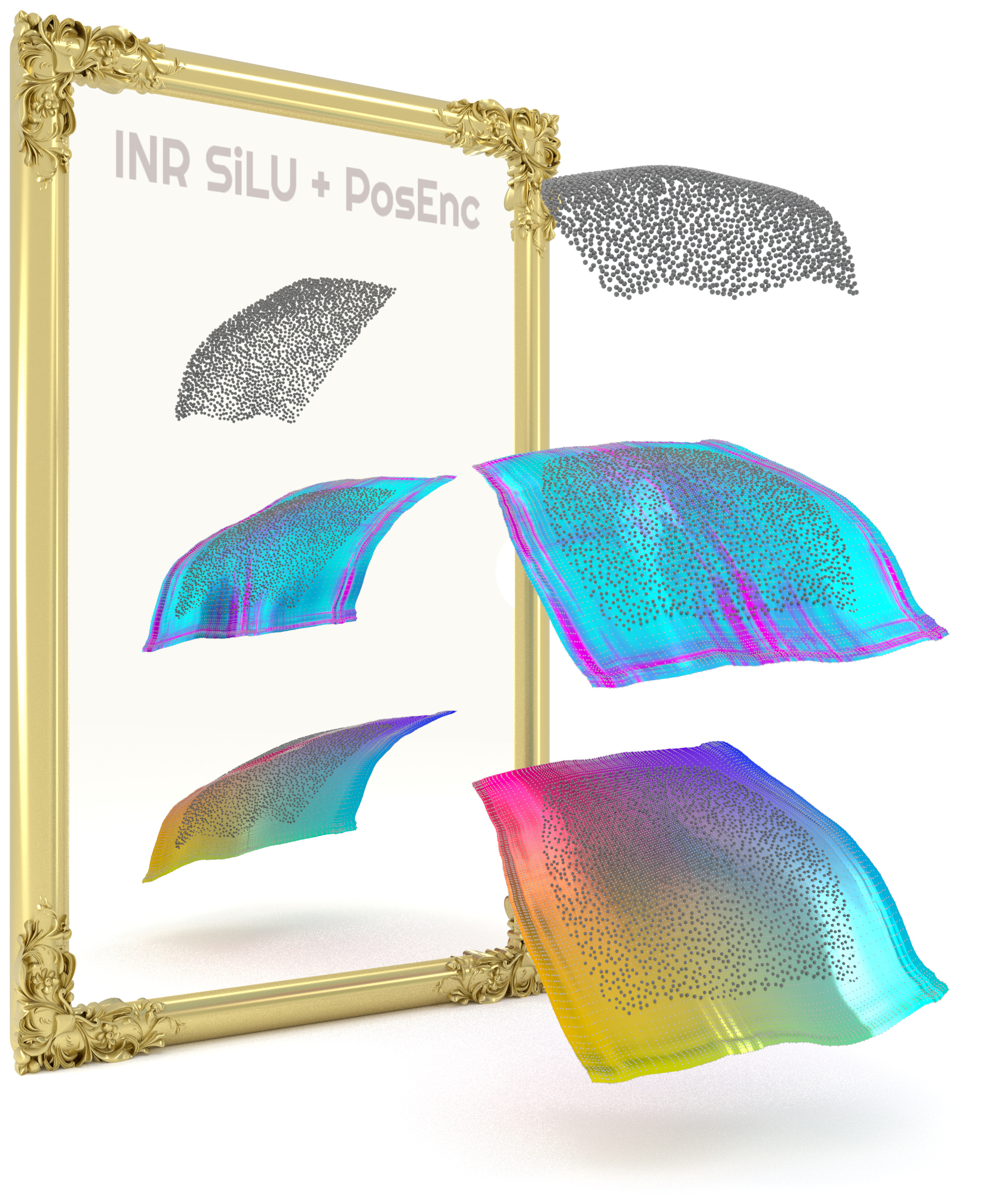}
      \hfill
   \includegraphics[width=0.24\linewidth]{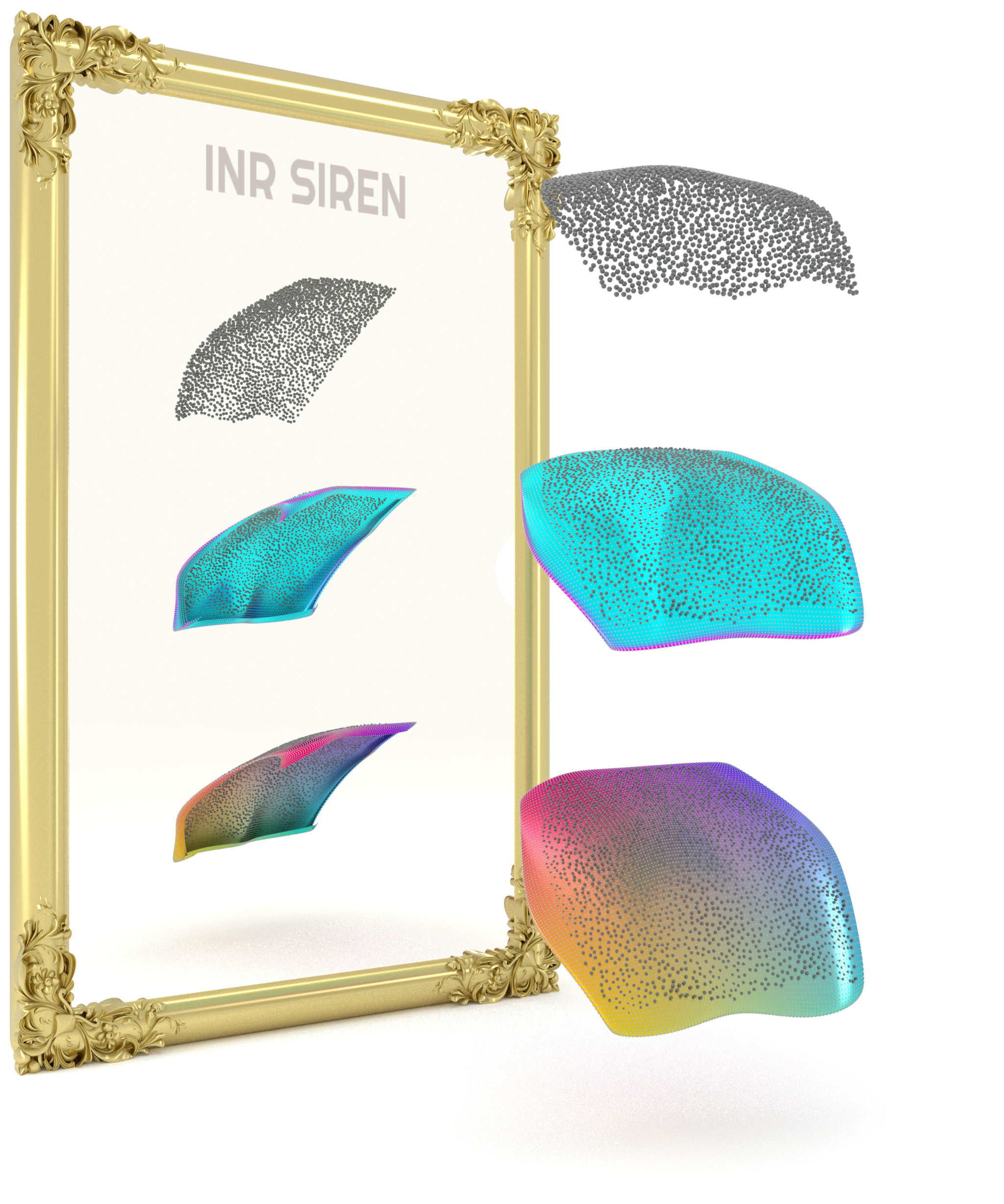}
\end{center}
    \caption{
    Inspiration items of our INR surface fitting. 
    UMAP~\cite{umap} (\textbf{1$^\mathrm{st}$ column, middle}) learns the underlying 2D manifold of 3D points along with the inverse mapping, but cannot capture its smooth analytic representation. 
    Early experiments with ReLU activations confirmed its low-frequency bias (\textbf{1$^\mathrm{st}$ column, bottom}) and piece-wise linearity. 
    SiLU~\cite{silu} (\textbf{2$^\mathrm{nd}$ column}) suffers from low-frequency bias, too. 
    We found that resolving it by adding positional encoding is challenging under our training protocol (\textbf{3$^\mathrm{rd}$ column}). 
    SIREN~\cite{siren} (\textbf{4$^\mathrm{th}$ column}) fits the data precisely, however, does not extrapolate well.
    A combination of the best parts of these ingredients led us to our INR surface fitting method, seen in Fig.~\ref{fig:inr:good} and discussed in Sec.~\ref{method:inr}.
    }
\label{fig:inr:ablations}
\end{figure*}

\section{Method}
\label{sec:method}

Contrary to the recent trend towards generic, end-to-end deep learning pipelines, we found it advantageous to split the reconstruction process into steps and only use neural methods where necessary.
Overall, our method consists of the following stages, \cf Fig.~\ref{fig:teaser}:
\begin{enumerate}[leftmargin=\parindent]
\item Partition the point cloud into clusters corresponding to the CAD model's topological faces. We rely on existing (pretrained) neural network methods for that step.
\item Fit an analytical surface primitive to each cluster. Here we use a hybrid approach: First, we test a set of prevalent geometric surface primitives that admit efficient closed-form fitting. We propose a novel fitting scheme for freeform surfaces based on neural representations.
\item Identify the effective area of each (finite) parametric surface and clip it, leaving enough margin to ensure intersections of adjacent surfaces.
\item Perform pairwise surface intersections to obtain a set of topologically plausible object edges. Using these edges, remove parts not supported by input points.
\item Perform pairwise edge intersection to identify a set of topological corners. Clip edges based on proximity to the remaining surface regions using these corners.
\end{enumerate}
As a result of applying Point2CAD, we obtain a CAD model in $B$-rep format, which includes analytical surfaces to represent the model's faces, compatible edges and corners, and adjacency matrices that encode topology.

A cornerstone of the proposed pipeline is a novel, efficient method for fitting the implicit neural representation of a freeform, parametric 2D-manifold surface in 3D to an unordered set of points. A particular difficulty in this context is that the surface must be extrapolated beyond the supporting points to enable robust surface intersections, an ability that tends to be challenging for neural surface estimators.
To that end, we design a novel freeform surface parameterization, drawing inspiration from recent advances in neural fields, positional encoding schemes, and efficient test-time optimization.
See Sec.~\ref{method:inr} for details.

While using neural methods, INR is a pure test-time optimization. 
We effectively side-step learning priors from training data except for the initial low-level segmentation. 
We argue that simple analytical procedures like surface fitting do not necessarily benefit from learned priors and that an over-reliance on data-driven learning may have hampered recent work on reverse engineering CAD models.
Moreover, separately predicting -- rather than constructing -- the topology does not guarantee that it is consistent with the geometry, \eg, edges could be instantiated such that they do not lie in a surface, see Fig.~\ref{fig:gallery}.
We empirically support this claim in our experiments, where we outperform purely learning-based methods like~\cite{guo2022complexgen}, which are inherently plagued by even slight differences between the training and test distributions.
When designing Point2CAD, we aimed for a generalizable scheme that uses offline learning where necessary -- in particular, for feature learning and segmentation, ill-posed tasks that are notoriously hard to engineer heuristically; but that does not replace well-understood analytical operations with less robust, data-driven feed-forward modeling. As a side effect, obviating parameter-heavy neural networks, where possible, also results in a computationally less demanding method.

\subsection{Parametrization of standard primitives}
The geometric primitives that our current implementation handles are planes, spheres, cylinders, and cones. 
The exact parameterizations are taken as in~\cite{sharma2020parsenet}, see Sec.~\ref{sec:primfitting} of sup.mat. 

We observed that utilizing the predicted surface type may harm the reconstruction since the prediction is not always correct. Therefore, we circumvent that step and instead rely on an exhaustive, data-driven comparison: we fit each primitive type (including INR) to point clusters and select the simplest model with the lowest reconstruction error. 

\subsection{Freeform surface parametrization with INR}
\label{method:inr}

In the context of the overall CAD reconstruction pipeline, this step should satisfy several requirements:
(1) resilience to noise in the input point cloud, which is inevitable with real 3D sensors; 
(2) support for inverse mapping, to enable traversal of the latent space;
(3) flexibility in interpolation mode, to ensure high data fidelity and avoid over-smoothing;
(4) strong regularization in extrapolation mode, required to construct smooth, plausible margins around the observed data points for the subsequent surface intersection;
(5) fast fitting with low computational cost. 

Existing manifold learning techniques that support the inverse transform, such as UMAP~\cite{umap}, could not be readily used for this task, as demonstrated in Fig.~\ref{fig:inr:ablations} (1$^\mathrm{st}$ column, middle). 
We thus developed a custom neural autoencoder~\cite{kramer1991nonlinear} with a single hidden layer, which is sufficient for simple non-linear transforms~\cite{hornik1989multilayer}.

We found that for our purposes, the choice of activation function is critical:
Besides causing a non-smooth, piece-wise linear approximation, ReLU activations~\cite{fukushima1975cognitron} exhibit a noticeable low-frequency bias~\cite{rahaman2019spectral}, as can be seen in Fig.~\ref{fig:inr:ablations} (1$^\mathrm{st}$ column, bottom). High-frequency undulations are over-smoothed, such that entire point cloud regions noticeably deviate from the fit on either side.
Replacing ReLU with SiLU~\cite{silu} enhances surface smoothness, but does not overcome the low-frequency bias (2$^\mathrm{nd}$ column). 
Adding positional encoding~\cite{nerf} in conjunction with SiLU resolves the issue, as one would expect from the literature, but provokes new artifacts, due to the fixed set of spatial frequencies (3$^\mathrm{rd}$ column).
Sinusoidal activations~\cite{siren} (4$^\mathrm{th}$ column) achieve excellent data fitting, but impose little regularization when not supported by data, and thus extrapolate poorly.

A combination of activations with desired properties within the same layer overcomes these individual problems: equipping some neurons with SiLU and the remaining ones with sinusoidal activations produces geometrically faithful fits that extrapolate smoothly, see Fig.~\ref{fig:inr:good}. Remarkably, this behavior emerges without any dedicated mechanisms to steer the collaboration between the two types of neurons.

\begin{figure}[t]
\begin{center}
\includegraphics[width=1.0\linewidth]{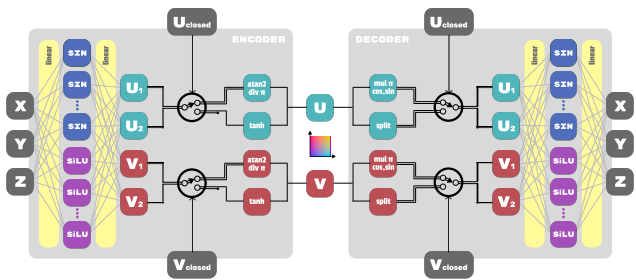}
\end{center}
\caption{
  Block scheme of our proposed INR for freeform surface fitting.
  We encode groups of 3D surface points into a latent 2D $uv$ space. We use a mixture of activations to achieve high fidelity of fitting and smooth extrapolation. Both open and closed surface fitting is supported via preconfigured routing of activations in the latent space. See Sec.~\ref{method:inr} for more details.
}
\label{fig:diagram}
\end{figure}

Fig.~\ref{fig:diagram} depicts a block diagram of the proposed autoencoder.
Each surface is autoencoded independently, by feeding batches of its 3D points ($X$,$Y$,$Z$) through a 1-layer MLP encoder and a corresponding decoder, with a 2D bottleneck corresponding to manifold coordinates ($u$,$v$). 
The 2D latent space is regularized to a unit square $[-1,1]\times[-1,1]$. Both closed, and open surfaces (independently in the $u$ and $v$ dimensions) are supported via pre-configured routing.

For each surface, the weights of a template INR are initialized randomly and optimized with standard mini-batch (or full-batch) descent. We run Adam~\cite{adam} for 1000 steps, with 50 steps warm-up of the learning rate, followed by a linear decay that reaches zero at the last step. 
The complete optimization takes only a few seconds on a single GPU, and multiple surfaces can be fitted in parallel.

For extrapolation and latent space traversal, we encode all cluster points into the latent $uv$ space and store the bounding box parameters along with the autoencoder.
We reset the corresponding axis range for surfaces with closed dimensions to $[-1,1]$.
To sample the extended surface with the margin, we extend the bounding box by $10\%$ in both dimensions and compute 3D surface points using the decoder.

\subsection{Topology reconstruction}

Following the instantiation of all individual surfaces, it becomes necessary to establish their boundaries, which amounts to finding the intersection curves between adjacent surfaces. Since this step is crucial for subsequent stages of CAD reconstruction, it is important to correctly identify and recover the intersection curves. This requirement again calls for deterministic geometric methods, as independent detection can hardly guarantee a set of complete edges consistent with the surfaces.

While it is theoretically possible to work out the surface intersections analytically, multiple freeform surfaces lead to extremely complicated, and potentially unstable calculations.
Hence, we opt for a generic solver and convert all surfaces to triangle meshes for this step to be accurately intersected with mature, numerically stable computational tools.
For each surface, the (infinite) geometric primitive is trimmed to form a margin of width $\epsilon$ around the supporting points to ensure enough space for the subsequent steps and then fed to a standard meshing algorithm by tessellation and then triangulation. 
We identify all intersecting surfaces for each meshed surface, compute the pairwise mesh intersections to obtain poly-line edges, and re-mesh the surfaces along those edges.
Similarly, we intersect adjacent edge poly-lines to obtain corner points and, in turn, use those corners to trim the edges (which still extend beyond the surfaces since they were generated from unclipped surfaces). See Sec.~\ref{sec:alg} of sup.mat. for a detailed algorithm.

Such a discretization step temporarily side-steps the $B$-rep format. However, it is possible to compute the true analytical intersections starting from the discrete approximations. This arguably simpler task is left for future work.

\begin{figure*}[t]
\begin{center}
\includegraphics[width=1.0\linewidth]{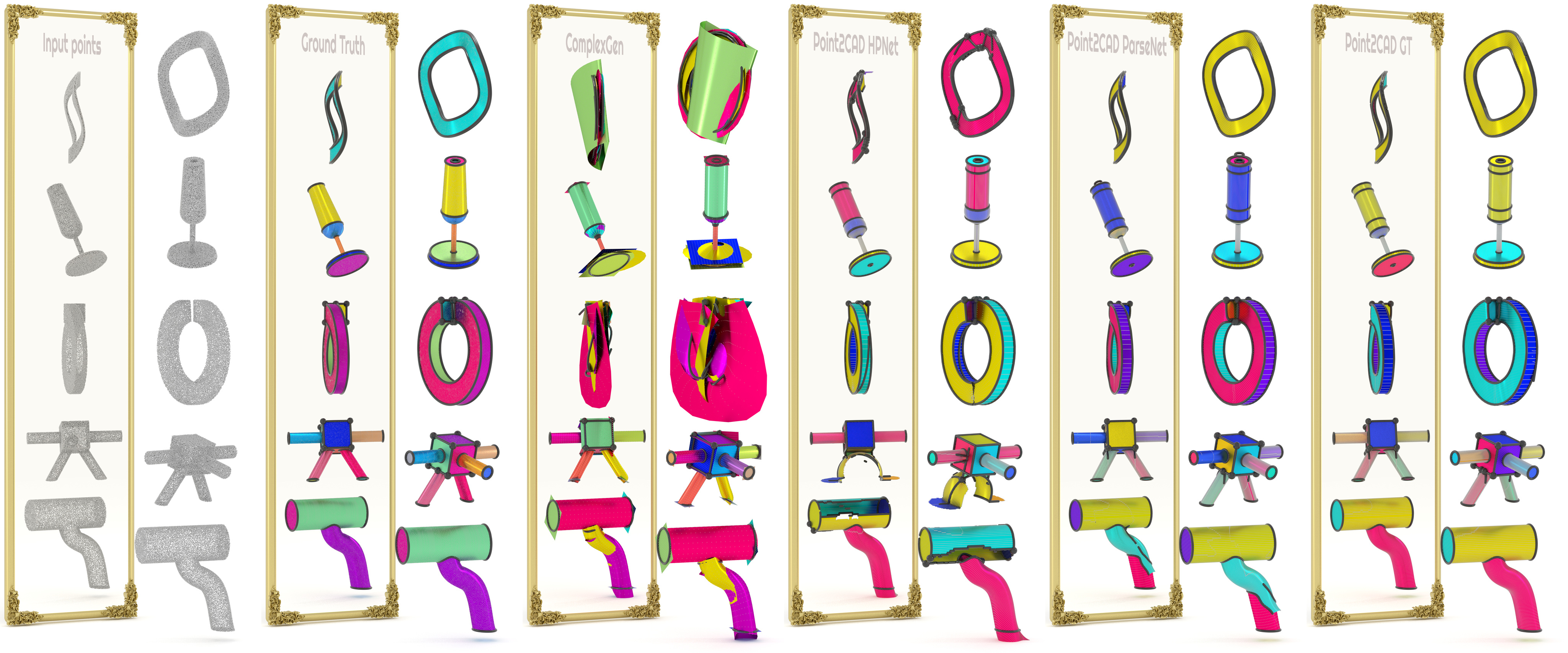}
\end{center}
\caption{
  The final results gallery, with different colors denoting different topological surfaces, and edges and corners depicted in black. 
  The first three, left to right: input point cloud, ground truth mesh, and reconstruction with ComplexGen~\cite{guo2022complexgen}. 
  The last three are the proposed Point2CAD method applied to different segmentation of the input points: with HPNet~\cite{yan2021hpnet}, ParSeNet~\cite{sharma2020parsenet}, and Ground Truth. 
  Our method reconstructs the ground truth geometry and topology from the ground truth segmentation nearly perfectly. 
  When applied on top of pretrained segmentation modules, it outperforms the competition by a high margin.
}
\label{fig:gallery}
\end{figure*}

\section{Experiments}
\label{sec:experiments}

The rather flexible constraints on the input to our method permit employing it in various setups. Such may include: 
\begin{enumerate}[leftmargin=\parindent]
\item[--] Evaluation of the ground truth point cloud clustering or segmentation as a way to quantify the contribution of point cloud sampling sparsity and sample noise on the reconstruction quality (aliased ``Point2CAD GT'');
\item[--] Usage of geometry fitting and topology extraction on top of any pretrained point cloud clustering or segmentation methods, such as ParSeNet~\cite{sharma2020parsenet} or HPNet~\cite{yan2021hpnet} (aliased ``Point2CAD $\langle$method$\rangle$'').
\end{enumerate}
Under both setups, we first individually evaluate the surface reconstruction quality of the most challenging freeform surfaces, which is one of the main contributions of our work. 
We then analyze the geometry and topology of the entire reconstructed CAD dataset. This enables us to provide an overall assessment of the effectiveness of our approach in reconstructing surfaces and CAD models from point clouds.

\subsection{Dataset}
To evaluate the performance of Point2CAD, we conduct experiments on the ABC dataset~\cite{koch2019abc}, a large-scale collection of CAD models ($\sim$1.000.000 models) widely used in geometric deep learning.
We use the same split as ParseNet, where each model contains at least one freeform surface, to facilitate a fair comparison with existing methods and to better demonstrate the feasibility of our approach.

\subsection{Evaluation metrics}
\begin{itemize}[leftmargin=\parindent]
\item[--] \textbf{Residual error} measures the discrepancy between the reconstructed surface and the corresponding ground truth surface obtained by Hungarian matching. 
Following the established protocol~\cite{sharma2020parsenet,yan2021hpnet}, we densely sample points on the ground truth surface and evaluate their correspondence with the reconstructed surface. Specifically, we compute the distance between each sampled point and its projected point on the reconstructed surface, which enables us to assess how well the fitting algorithm has captured the underlying geometry of the original surface. The residual error for $i$-th surface is:
\begin{equation}
Err_i =\frac{1}{N_{i}}\sum_{\mathbf{s}_{i,k}\in S_{i}^{\text{gt}}} d(\mathbf{s}_{i,k}, S_{i}^r),
\end{equation}
where $N_{i}$ is the number of sampled points, $\mathbf{s}_{i,k}$ is the $k$-th sampled point on the $i$-th ground truth surface $S_{i}^{\text{gt}}$, and its corresponding reconstructed surface is $S_{i}^r$. The residual error $Err_i$ is computed as the average Euclidean distance $d(\cdot, \cdot)$ between the sampled points and their corresponding projections on the reconstructed surface. The residual error for $i$-th surface is
a simple average can provide an overall assessment of a CAD model, $Err=\frac{1}{M}\sum_{i=1}^{M}Err_i$.

\item[--] \textbf{P-coverage}. It quantifies the proportion of input point cloud covered by the generated surface:
\begin{equation}
Pcov  = \frac{1}{|P|}\sum_{k=1}^{|P|} \mathbb{I}\{d(p_k, S^r) \leq r\},
\end{equation}
where  $|P|$ is the number of points in the entire input point set $P$ and $S$ is all CAD reconstruction surfaces. $r$ is a distance threshold, set as $0.01$ following previous works.

\item[--] \textbf{Surface precision, recall, F-score}.
We adopted several evaluation metrics introduced in ComplexGen~\cite{guo2022complexgen} to assess the efficacy of our method in generating reconstructed surfaces. 
A surface is considered a true positive if the distance between it and its closest ground truth surface is below a certain threshold $\theta$, which serves as a measure of accuracy for the method in a geometric sense.

\item[--] \textbf{Edge precision, recall, F-score}.
The effectiveness of the proposed method can also be assessed from a topological perspective, by computing the precision, recall, and F-score on the edges resulting from pairwise surface intersections. This evaluation strategy provides a reliable measure of the method's ability to capture the topology of the reconstructed surfaces.

\item[--] \textbf{Corner precision, recall, F-score}.
Likewise, an accurate estimation of corner positions resulting from edge intersections is an essential indicator of overall accuracy.

\item[--] \textbf{Chamfer distance}. 
This metric assesses the accuracy of the reconstructed CAD model as a whole, without distinguishing between individual surfaces. To that end, we sample the entire surface of both the reconstructed model and the ground truth object and calculate the chamfer distance between the two point sets.
\end{itemize}

\subsection{Evaluation on freeform surfaces}
Both ComplexGen~\cite{guo2022complexgen} and ParseNet~\cite{sharma2020parsenet} have successfully demonstrated the prediction of the freeform surfaces. 
Both papers employ a custom parameterization that represents any freeform surface as a structured grid of 20x20 control points, regardless of the underlying scale. 

\begin{table}[t!]
\centering
\caption{
Comparison of surface fitting.
ComplexGen uses a transformer decoder to convert a surface latent code to a 20x20 control-points grid. 
ParseNet employs a neural network for the same kind of output. 
Our method utilizes the proposed INR surface format.
}
\resizebox{0.9\linewidth}{!}{
\begin{tabular}{@{}lcccc@{}}
\toprule
& \multicolumn{2}{c}{Open surfaces} & \multicolumn{2}{c}{Closed surfaces}   \\
\cmidrule(lr){2-3} \cmidrule(lr){4-5} 
& Res-err~$\downarrow$ & P-cover~$\uparrow$ & Res-err~$\downarrow$ & P-cover~$\uparrow$ \\
\midrule
ComplexGen & $0.021$ & $0.938$ & $0.023$ & $0.900$ \\
ParseNet & $0.006$ & $0.930$ & $0.008$ & $0.902$\\
Ours & $\mathbf{0.002}$ & $\mathbf{0.999}$ & $\mathbf{0.003}$ & $\mathbf{0.998}$\\
\bottomrule
\end{tabular}
 }
\label{tab:geo:spline}
\end{table}

\begin{table}[t]
\centering
\caption{Ablation of INR design performance with and w/o noise added to the input points. 
Our method demonstrates favorable properties as measured by residual error and P-coverage.
}
\resizebox{\linewidth}{!}{
\begin{tabular}{@{}crcccc@{}}
\toprule
& & \multicolumn{2}{c}{Open surfaces} & \multicolumn{2}{c}{Closed surfaces}   \\
\cmidrule(lr){3-4} \cmidrule(lr){5-6} 
Noise
& Activation & Res-err~$\downarrow$ & P-cover~$\uparrow$ & Res-err~$\downarrow$ & P-cover~$\uparrow$ \\
\midrule
\multirow{5}{*}{\rotatebox[origin=c]{90}{$\sigma=0.05$}}  
& ReLU & $0.016$ & $0.519$ & $0.018$ & $0.509$ \\
& SiLU & $0.024$ &$0.522$ & $0.030$ & $0.543$\\
& SiLU+PosEnc & $0.014$ &$0.561$& $0.015$ & $0.516$\\
& SIREN &$0.012$ &$0.524$ & $0.014$ & $0.521$\\
& Ours & $\mathbf{0.011}$ & $\mathbf{0.585}$ & $\mathbf{0.013}$ & $\mathbf{0.581}$\\
\midrule
\multirow{5}{*}{\rotatebox[origin=c]{90}{No noise}}  
& ReLU & $0.006$ & $0.986$ & $0.008$ & $0.981$\\
& SiLU & $0.020$ & $0.905$ & $0.023$ & $0.887$\\
& SiLU+PosEnc & $0.004$ & $0.992$ & $0.007$ & $0.982$\\
& SIREN & $0.003$ & $0.996$ & $0.006$ & $0.990$\\
& Ours & $\mathbf{0.002}$ & $\mathbf{0.999}$ & $\mathbf{0.003}$ & $\mathbf{0.998}$\\
\bottomrule
\end{tabular}
}
\label{tab:spline_ablation}
\end{table}

\begin{table}[t]
\centering
\caption{Ablation of INR under surface margin extrapolation by a fraction $c$ of model diameter. Point2CAD uses $c=10\%$.}
\resizebox{0.6\linewidth}{!}{
\begin{tabular}{@{}lccc@{}}
\toprule
Res-err~$\downarrow$ & $c=0\%$ & \fbox{$c=10\%$}  & $c=20\%$ \\
\midrule
SiLU & $0.0582$ & $0.0565$ &  $0.0714$ \\
SIREN & $0.0195$ & $0.0376$ & $0.1205$ \\
Ours & $\mathbf{0.0052}$ & $\mathbf{0.0156}$ & $\mathbf{0.0474}$  \\
\bottomrule
\end{tabular}
} 
\label{tab:apron}
\end{table}
\begin{table}[t]
\centering
\caption{Geometric evaluation of reconstructed CADs.
Segmentation denotes the method used for point cloud clustering.
``GT'' stands for oracle ground truth segmentation, which is also an upper bound of the performance of our method. }
\resizebox{0.9\linewidth}{!}{
\begin{tabular}{@{}lcccc@{}}
\toprule
& Segmentation & Res-err~$\downarrow$  & P-cover~$\uparrow$ & Chamfer~$\downarrow$ \\
\midrule
ComplexGen & N/A & $0.020$ & $\mathbf{0.950}$ & $0.042$ \\
Point2CAD & ParseNet & $\mathbf{0.018}$ & $0.942$ & $\mathbf{0.017}$\\
Point2CAD & HPNet & $0.020$ & $0.937$ & $0.018$ \\
\midrule
Point2CAD & GT & $0.011$ & $0.968$ & $0.016$\\
\bottomrule
\end{tabular}
} 
\label{tab:geo:all}
\end{table}

To allow for an apples-to-apples comparison, we seek to obtain the results of all three methods on distinct point cloud clusters. 
While this is straightforward with ParseNet and our method, ComplexGen does not operate on individual clusters but instead requires the entire point cloud at once and outputs all $B$-rep elements at once, too.
Thus, to obtain ComplexGen predictions for a particular cluster, we run it with the entire point cloud input, find the nearest predicted surface to the queried cluster out of all the predictions, and perform the surface-to-surface evaluation on that pair.

We repeat this procedure for a subset of freeform surfaces of the ABC dataset and average the geometry metrics; see Tab.~\ref{tab:geo:spline} for quantitative evaluation. 
As can be seen, our INR fitting method generates freeform surfaces that represent the underlying points more faithfully than prior parameterizations. 
Qualitative results of Point2CAD freeform surface fitting confirm the qualitative study, which can be seen in Fig.~\ref{fig:inr:good}, and the side-by-side comparison is in Fig.~\ref{fig:gallery}. 

Additionally, we studied the behavior of different INR activations under noise-induced corruption of the input point clouds (Tab.~\ref{tab:spline_ablation}) and surface margin extrapolation (Tab.~\ref{tab:apron}), a crucial property of our pipeline.
For the latter, given a point cloud patch of a freeform surface, we select a subset of points from the center of the patch for fitting, leaving a margin of $c\%$ for extrapolation evaluation.
Our INR design consistently performs well across various settings.

\subsection{Reconstructed CAD evaluation}
We conduct an evaluation of Point2CAD in terms of the reconstruction quality for surfaces, edges, and corners. We compare our results to those obtained with ComplexGen~\cite{guo2022complexgen}, which is also capable of generating all those geometric elements and their topological relationships. 
However, it should be noted that the models generated by ComplexGen are often invalid, in the sense that the predicted geometry and topology are inconsistent. We are not aware of an easy way to rectify this issue. A possible solution could be only to utilize the surfaces generated by ComplexGen and apply our method to refit and intersect them.

\begin{table*}[t]
\centering
\caption{Evaluation on CAD Reconstruction of \textit{Surfaces}, \textit{Edges} and \textit{Corners} from the aspects of accuracy and completeness. The surface evaluation assesses the reconstruction performance in geometric terms, while these metrics on edges and corners reflect the performance of reconstruction in terms of topological properties. Given that the distance calculation is based on sampling, we adopt various threshold values for evaluating surfaces, edges and corners. Specifically, the threshold for a single-point corner is expected to be smaller than that of edges, while the threshold for edges should be smaller than that of surfaces, which aligns with the intuitive expectations in CAD evaluation.
}
\resizebox{\linewidth}{!}{
\begin{tabular}{@{}lcccccccccc@{}}
\toprule
\multicolumn{2}{c}{Surfaces} & \multicolumn{3}{c}{$\theta_{\text{surface}}=0.08$} & \multicolumn{3}{c}{$\theta_{\text{surface}}=0.06$} & \multicolumn{3}{c}{$\theta_{\text{surface}}=0.03$}  \\
\cmidrule(lr){3-5} \cmidrule(lr){6-8} \cmidrule(lr){9-11}
Method & Segmentation & precision~$\uparrow$ & recall~$\uparrow$ & F-score~$\uparrow$ & precision~$\uparrow$ & recall~$\uparrow$ & F-score~$\uparrow$ & precision~$\uparrow$ & recall~$\uparrow$ & F-score~$\uparrow$ \\
\midrule
ComplexGen & N/A & $0.732$ & $0.732$ & $0.731$ & $0.633$ & $0.641$ & $0.637$ & $0.370$ & $0.388$ & $0.379$ \\
Point2CAD & ParseNet & $0.817$ & $\mathbf{0.750}$ & $0.782$ & $0.764$ & $0.693$ & $0.727$ & $0.624$ & $0.562$ & $0.591$ \\
Point2CAD & HPNet & $\mathbf{0.836}$ & $0.745$ & $\mathbf{0.788}$ & $\mathbf{0.796}$ & $\mathbf{0.695}$ & $\mathbf{0.742}$ & $\mathbf{0.697}$ & $\mathbf{0.593}$ & $\mathbf{0.641}$\\
\midrule
Point2CAD & GT &$0.976$ & $0.920$ & $0.947$ & $0.962$ & $0.898$ & $0.929$ & $0.873$ & $0.801$ & $0.836$\\
\midrule[0.1em]
%
\multicolumn{2}{c}{Edges} & \multicolumn{3}{c}{$\theta_{\text{edge}}=0.05$} & \multicolumn{3}{c}{$\theta_{\text{edge}}=0.03$} & \multicolumn{3}{c}{$\theta_{\text{edge}}=0.02$}  \\
\cmidrule(lr){3-5} \cmidrule(lr){6-8} \cmidrule(lr){9-11}
Method & Segmentation & precision~$\uparrow$ & recall~$\uparrow$ & F-score~$\uparrow$ & precision~$\uparrow$ & recall~$\uparrow$ & F-score~$\uparrow$ & precision~$\uparrow$ & recall~$\uparrow$ & F-score~$\uparrow$ \\
\midrule
ComplexGen & N/A & $0.620$ & $0.576$ & $0.597$ & $0.421$ & $0.397$ & $0.409$ & $0.291$ & $0.279$ & $0.285$ \\
Point2CAD & ParseNet & $0.636$ & $\mathbf{0.596}$ & $\mathbf{0.615}$ & $0.523$ & $\mathbf{0.486}$ & $\mathbf{0.504}$ & $0.440$ & $\mathbf{0.414}$ & $\mathbf{0.427}$ \\
Point2CAD & HPNet & $\mathbf{0.637}$ & $0.586$ & $0.611$ & $\mathbf{0.526}$ & $0.473$ & $0.498$ & $\mathbf{0.449}$ & $0.397$ & $0.421$ \\
\midrule
Point2CAD & GT & $0.863$ & $0.774$ & $0.816$ & $0.766$ & $0.673$ & $0.717$ & $0.686$ & $0.597$ & $0.639$\\
\midrule[0.1em]
%
\multicolumn{2}{c}{Corners} & \multicolumn{3}{c}{$\theta_{\text{corner}}=0.03$} & \multicolumn{3}{c}{$\theta_{\text{corner}}=0.02$} & \multicolumn{3}{c}{$\theta_{\text{corner}}=0.01$}  \\
\cmidrule(lr){3-5} \cmidrule(lr){6-8} \cmidrule(lr){9-11}
Method & Segmentation & precision~$\uparrow$ & recall~$\uparrow$ & F-score~$\uparrow$ & precision~$\uparrow$ & recall~$\uparrow$ & F-score~$\uparrow$ & precision~$\uparrow$ & recall~$\uparrow$ & F-score~$\uparrow$ \\
\midrule
ComplexGen & N/A & $\mathbf{0.667}$ & $0.633$ & $0.650$ & $0.483$ & $0.454$ & $0.468$ & $0.217$ & $0.203$ & $0.210$ \\
Point2CAD & ParseNet & $0.661$ & $\mathbf{0.641}$ & $\mathbf{0.651}$ & $\mathbf{0.553}$ & $\mathbf{0.529}$ & $\mathbf{0.541}$ & $\mathbf{0.398}$ & $\mathbf{0.392}$ & $\mathbf{0.395}$ \\
Point2CAD & HPNet & $0.646$ & $0.613$ & $0.629$ & $0.549$ & $0.521$ & $0.535$ & $0.389$ & $0.386$ & $0.388$\\
\midrule
Point2CAD & GT & $0.780$ & $0.696$ & $0.736$ & $0.704$ & $0.662$ & $0.661$ & $0.581$ & $0.515$ & $0.546$\\
\bottomrule
\end{tabular}

} 
\label{tab:topo:allthree}
\end{table*}

As seen in Tab.~\ref{tab:geo:all}, Point2CAD outperforms ComplexGen in most metrics. Furthermore, Point2CAD achieves excellent results when fed with ground truth segmentations, indicating that it will likely further improve as better point cloud segmentation engines become available.
We also observed that our P-coverage results were slightly inferior to ComplexGen's.
This can be attributed to the fact that, in the final step of our reconstruction, we trim the margins of the individual surfaces to establish a clear and well-defined boundary. 
As a result, the coverage of our reconstructed CAD models is somewhat reduced compared to ComplexGen, which does not perform this step and typically generates more surfaces than the ground truth has. The presence of numerous spurious surfaces naturally inflates the P-coverage. This limitation of the metric is confirmed by the fact that we obtain higher P-coverage if we evaluate Point2CAD without clipping the surfaces at the edges (Point2CAD+ParseNet: $0.956$, Point2CAD+HPNet: $0.953$, Point2CAD+GT: $0.971$).

We also employ several standard evaluation metrics, including precision, recall, and F-score, to measure the accuracy and completeness of the surfaces after the overall reconstruction of CAD under different distance thresholds $\theta_{\text{surface}}$, see Tab.~\ref{tab:topo:allthree}~(top).
The results suggest that Point2CAD with HPNet segmentation backbone achieves the highest overall performance, while reconstruction based on the ground truth point cloud segmentation provides the best results, as expected.

We utilize the precision, recall, and F-score of the reconstructed curves and corners as metrics to evaluate the topological reconstruction performance; results are shown in Tab.~\ref{tab:topo:allthree}~(middle, bottom).

\section{Conclusion}
\label{sec:conclusion}
We present Point2CAD, a comprehensive and versatile approach for reverse engineering CAD models from point clouds, which can yield various types of surfaces and topologically correct reconstructions. 
The proposed method segments point clouds with a pre-trained segmentation backbone but then employs learning-free optimization methods to fit geometric primitives, including freeform surfaces that are optimized with a novel neural surface estimation scheme. 
The recovered surfaces are analytically intersected to obtain the edges and corners of the model.
Empirically, our proposed method qualitatively and quantitatively outperforms prior art on the popular ABC dataset of CAD models. In our view, these results support our hypothesis that the main role of machine learning in the context of CAD reconstruction is actually segmentation, whereas subsequent steps are adequately solved by classical geometric operations and may not always require learning.

\noindent\textbf{Limitations} 
While our method offers a solution to CAD model reconstruction, it is subject to certain limitations.
Notably, our method is sensitive to the quality of the point cloud segmentation. 
A number of outliers in the prediction of points belonging to the same surface can adversely impact the subsequent surface fitting and topology reconstruction. 
Despite the fact that our method is not end-to-end-trainable, the main guideline in our design is to divide the pipeline into isolated, accountable steps and only employ neural methods where needed.
While this runs against a puristic deep learning spirit, our modular pipeline outperforms the end-to-end learned solutions. 
We see this as an important message, although we do not rule out that it could serve as a foundation for future end-to-end works.

\noindent\textbf{Future work} 
One promising direction would be to exploit the proposed INR fitting scheme in a feedback loop of any other algorithm to drive point cloud partitioning into clusters, thus lifting the dependence of the reconstruction pipeline on the quality of point cloud segmentation.
Differently, bringing back learned priors and the end-to-end property should make it possible to further improve the results.

{
    \small
    \bibliographystyle{ieeenat_fullname}
    \bibliography{main}
}
\clearpage
\setcounter{page}{1}
\maketitlesupplementary

\section{Reproducibility}
\label{sec:reproducibility}

Please refer to the attached supplementary zip file for the source code of our method. See \texttt{README.md} for instructions on setting up the environment and running the code. 

\section{Notation}
\label{sec:notation}

Throughout the paper, we use several terms and concepts as follows. 
A \textbf{CAD model} represents a complex object in a format that permits editing in CAD software.
\textbf{Boundary representation (B-REP)} is one such representation; it involves analytic \textbf{surfaces}, \textbf{edges}, and \textbf{corners}\footnote{
Related terms in literature: ``surfaces'' := ``faces'', ``patches''; ``edges'' := ``curves'', ``contours''; ``corners'' := ``points'', ``vertices'', ``endpoints''.
}%
, and explicit topological relationships between them.
Typically, such relationships take the form of ``an edge is an intersection of two surfaces'' and ``a corner is an intersection of two edges'', stated as adjacency matrices.
The manifolds of surfaces and edges have intrinsic dimensionality of 2 and 1 respectively.
\textbf{Point cloud} refers to the unordered set of 3D points, obtained by scanning the surface of a real world object or by sampling a CAD model for simulation purposes. 
\textbf{Semantic classes} are the geometric types that we consider for representing CAD model surfaces. 
These typically include a plane, sphere, cylinder, cone, and freeform (spline) surfaces.
\textbf{Segmentation} refers to assigning a semantic class to each point of the point cloud.
\textbf{Clusters} are sets of points belonging to the same semantic class and instance; they group points belonging to distinct surfaces. 
\textbf{Clustering} information can be a byproduct of segmentation, an algorithmic stage, or given as ground truth.
\textbf{Primitives} are inherently low-dimensional parametric models representing analytical shapes of semantic classes, and that can be used to fit clusters with least squares. 



\section{Primitive parameterization and fitting}
\label{sec:primfitting}
Our method can handle several primitive types, including planes, spheres, cylinders, and cones~\cite{li2019supervised,geometrictools}. 

\paragraph{Plane}
In the context of 3D geometry, a plane can be represented by a vector $\mathbf{n}\in\mathbb{R}^3$, which is the unit normal to the plane, and a scalar $d$ that determines the distance of the plane from the origin. %
Mathematically, a plane can be denoted by a tuple $(\mathbf{n}, d)$, where $\mathbf{n}$ satisfies $|\mathbf{n}|=1$. Any point $\mathbf{p}\in\mathbb{R}^3$ lies on the plane if and only if $\mathbf{n}^\mathrm{T}\mathbf{p}=d$.

The following steps can be employed to perform plane fitting to a collection of 3D points:

\begin{enumerate}
    \item Calculate the centroid of the points:
$ \mathbf{c} = \frac{1}{n} \sum_{i=1}^{n} \mathbf{p_i} $;

\item Compute the matrix:
$ \mathbf{M} = \frac{1}{n} \sum_{i=1}^{n} (\mathbf{p_i} - \mathbf{c}) (\mathbf{p_i} - \mathbf{c})^T $;

\item Calculate the eigenvalues and eigenvectors of $\mathbf{M}$.
Select the eigenvector corresponding to the smallest eigenvalue as the normal vector $\mathbf{n}$.
Compute the distance d using the normal vector $\mathbf{n}$ and the centroid c:
$ d = \mathbf{n}^T \mathbf{c} $.
\end{enumerate}

\paragraph{Sphere}
A sphere can be represented by a tuple $\left(\mathbf{c}, r\right)$, where $\mathbf{c} \in \mathbb{R}^3$ denotes the center of the sphere and $r \in \mathbb{R}$ denotes its radius. Any point $\mathbf{p}\in\mathbb{R}^3$ lies on the sphere if and only if $||p-\mathbf{c}||=r$. The error function to be minimized is as follows: 
\begin{align}
E\left(\mathbf{c},r\right)=\sum_{i=1}^M \left(\|\mathbf{p}_i-\mathbf{c}\|-r\right)^2.
\end{align}
Solving $\frac{\partial E}{\partial r}=0$  for $r$ yields
\begin{align}
r = \frac{1}{M} \sum_{i=1}^{M} \|\mathbf{p}_i-\mathbf{c}\|.
\end{align}
Solving $\frac{\partial E}{\partial \mathbf{c}}=0$ for $\mathbf{c}$ gives
\begin{align}
\mathbf{c}=\frac{1}{M}\sum_{i=1}^M\mathbf{p}_i+r\cdot\frac{1}{M}\sum_{i=1}^M \frac{\partial(\|\mathbf{p}_i-\mathbf{c}\|)}{\partial \mathbf{c}}.
\end{align}
A fixed point iteration can solve the equations and subsequently determine the parameters $(\mathbf{c}, r)$.

\paragraph{Cylinder}
An infinite cylinder $(\mathbf{a}, \mathbf{c}, r)$ is characterized by a point $\mathbf{c}$, a unit-length direction vector $\mathbf{a}$ that defines its axis and a radius $r$. 
Any point $\mathbf{p}\in\mathbb{R}^3$ lies on the cylinder if and only if $\left(\mathbf{p}-\mathbf{c}\right)^T\left(\mathbf{I}-\mathbf{a}\mathbf{a}^T\right)\left(\mathbf{p}-\mathbf{c}\right)=r$
The error function to be minimized is
\begin{align}
E\left(\mathbf{a}, \mathbf{c}, r\right)\!=\!\sum_{i=1}^M\!\left(\left(\mathbf{p}_i\!-\!\mathbf{c}\right)^T\!\!\left(\mathbf{I}\!-\!\mathbf{a}\mathbf{a}^T\right)\!\left(\mathbf{p}_i\!-\!\mathbf{c}\right)-r^2\right)^2,
\end{align}
where $\mathbf{I}$ is the identity matrix. 
Solving $\frac{\partial E}{\partial r^2}=0$ leads to 
\begin{align}
r=\left(\frac{1}{M}\sum_{i=1}^M(\mathbf{c}-\mathbf{p}_i)^T(\mathbf{I}-\mathbf{a}\mathbf{a}^T)(\mathbf{c}-\mathbf{p}_i)\right)^{\frac{1}{2}}.
\label{eq:cylinder:r}
\end{align}
By solving $\frac{\partial E}{\partial \mathbf{c}}=0$, we can obtain an equation for $\mathbf{c}$:
\begin{align}
\mathbf{c}=\frac{\tilde{\mathbf{A}}^{-1}\mathbf{A}}{\operatorname{tr}(\hat{\mathbf{A}}\mathbf{A})}\left(\frac{1}{M}\sum_{i=1}^M\left(\mathbf{p}_i^T\tilde{\mathbf{A}}\mathbf{p}_i\right)\mathbf{p}_i\right),
\label{eq:cylinder:c}
\end{align}
where 
\begin{align}
\mathbf{A}=\tilde{\mathbf{A}}\left(\frac{1}{M}\sum_{i=1}^M \mathbf{p}_i\mathbf{p}_i^T\right)\tilde{\mathbf{A}},
\end{align}
and $\hat{\mathbf{A}}=\mathbf{S}\mathbf{A}\mathbf{S}^T$, $\tilde{\mathbf{A}} = \mathbf{I}-\mathbf{a}\mathbf{a}^T$, $\mathbf{S}$ is the skew symmetric matrix of $\mathbf{a}$.
Putting them back into the error function yields:
\begin{equation}
\begin{aligned}
G\left(\mathbf{a}\right)&
=\frac{1}{M}\sum_{i=1}^M \biggl[\mathbf{p}_i^T\tilde{\mathbf{A}}\mathbf{p}_i-\frac{1}{M}\sum_{j=1}^M\mathbf{p}_j^T\tilde{\mathbf{A}}\mathbf{p}_j \, - \\
&-\frac{2\mathbf{p}_i^T\hat{\mathbf{A}}}{\operatorname{Tr}\left(\hat{\mathbf{A}}\mathbf{A}\right)}\left(\frac{1}{M}\sum_{j=1}^M\left(\mathbf{p}_j^T\tilde{\mathbf{A}}\mathbf{p}_j\right)\mathbf{p}_j\right)\biggr]^2 .
\end{aligned}
\end{equation}
A Powell optimizer~\cite{powell1964efficient} is then employed to locate the global minimum of the function $G(\mathbf{a})$. Once $\mathbf{a}$ has been determined, the center $\mathbf{c}$ and radius $r$ of the corresponding circle can be easily obtained through Eq.~\ref{eq:cylinder:r} and \ref{eq:cylinder:c}.

\paragraph{Cone}
We parameterize an infinite cone using the set of parameters $(\mathbf{v}, \mathbf{a}, \theta)$, where $\mathbf{v}\in \mathbb{R}^3$ denotes the apex point, $\mathbf{a}\in \mathbb{R}^3$ denotes a unit axis direction vector, and $\theta \in (0, \pi/2)$ represents half the angle of the cone.
Any point $\mathbf{p}$ that lies on it satisfies
$\mathbf{a}\cdot\frac{\mathbf{p}-\mathbf{v}}{||\mathbf{p}-\mathbf{v}||}=\cos(\theta)$, which can be written in a quadratic form as
$(\mathbf{p}-\mathbf{v})^T(\cos(\theta)^2\mathbf{I}-\mathbf{a}\mathbf{a}^T)(\mathbf{p}-\mathbf{v})=0$. Thus, the error function can be defined as follows:
\begin{align}
\!\!E\!\left(\mathbf{v}, \!\mathbf{a}, \!\theta\right)\!=\!\sum_{i=1}^M\!\left(\!\left(\mathbf{p}_i\!-\!\mathbf{v}\right)^T\!\!\left(\!\cos\left(\theta\right)^2\!\mathbf{I}\!-\!\mathbf{a}\mathbf{a}^T\!\right)\!\left(\mathbf{p}_i\!-\!\mathbf{v}\right)\!\right)^2\!\!.
\end{align}
This least-square problem can be solved efficiently with the Levenberg-Marquardt~\cite{ranganathan2004levenberg} method.

\section{Algorithm of topological reconstruction}
\label{sec:alg}
We present a pseudocode-based methodology for reconstructing topological structures derived from $M$ surfaces fitted with potentially infinite primitives, forming a set $\{\mathbf{S}^0_k\}_M$. 
We denote the stage of surface processing with superscript.
We trim each primitive to form a margin of width $\epsilon$ around the input points, then employ tessellation and triangulation meshing algorithm. 
As a result, we obtain a set of finite extended surfaces denoted as $\{\mathbf{S}^1_k\}_M$. 
We generate poly-line edges $\{\mathbf{E}^1_k\}_K$ by identifying intersecting surfaces and computing pairwise intersections. And we trim the surfaces by the edges, thus obtaining $\{\mathbf{S}^2_k\}_M$.
Similarly, we intersect adjacent poly-line edges to obtain corner points $\{\mathbf{C}_k\}_L$, subsequently trimming edges accordingly to get final edges $\{\mathbf{E}^2_k\}_K$. See Alg.~\ref{alg:topo}.

\begin{algorithm}[t]
\caption{
Point2CAD model reconstruction
}
\label{alg:topo}
\begin{algorithmic}[1]
\Statex \textbf{Input:} $M$ surfaces $\{\mathbf{S}^0_k\}_M$ obtained from fitting primitives to point clusters $\{\mathbf{P}_k\}_M$
\Statex \textbf{Output:} a CAD model with $M$ surfaces $\{\mathbf{S}^2_k\}_M$, $K$ edges $\{\mathbf{E}^2_k\}_K$ and $L$ corners $\{\mathbf{C}_k\}_L$
\For{$i \in 1..M$}
\State Trim $\mathbf{S}^0_i$ by $\epsilon$ to input points:
$\mathbf{S}^1_i = \operatorname{Trim}(\mathbf{S}^0_i|\mathbf{P}_i,\epsilon)$
\EndFor
\For{$i \in 1..M$}
\State get edges on $\mathbf{S}^1_i$: $\{\mathbf{E}^1_{i,r}\}_{R_i}=\{\mathbf{S}^1_j\}_{j\neq i}\cap \mathbf{S}^1_i$
\State trim $\mathbf{S}^1_i$ by edges:
$\mathbf{S}^2_i=
\operatorname{Trim}(\mathbf{S}^1_i|\{\mathbf{E}^1_{i,r}\}_{R_i})$
\EndFor
\For{each pair $(\mathbf{E}_p, \mathbf{E}_q)$ in intersection edges}
\State obtain the corners $\mathbf{C}_{pq}= \mathbf{E}_p \cap \mathbf{E}_q$
\EndFor
\For{$i \in 1..L$}
\If{any $\mathbf{C}_s\in \mathbf{E}_i$}
\State trim it by the corners: $\mathbf{E}^2_i=\operatorname{Trim}(\mathbf{E}^1_i|\mathbf{C}_s)$
\EndIf
\EndFor
\end{algorithmic}
\end{algorithm}

Two distinct cases exist of using the $\operatorname{Trim}$ operation on surfaces. In the first case, trimming is performed based on a distance threshold, retaining only the portion of the infinite primitives near the input points. 
In the second case, we trim the surfaces by considering the intersection of their triangle-mesh representations with edges. 
We employ connected component analysis, whereby a pair of faces is considered connected if a path exists between them that does not cross an edge obtained through the surface intersection.
We then discard whole connected components based on the distance of their members to the original point cloud.
The $\operatorname{Trim}$ operation on edges involves a similar subdivision of the edge into connected segments by corners and retaining segments close to the input points.

\end{document}